\documentclass[journal]{IEEEtran}
\ifCLASSINFOpdf
\else
\fi
\usepackage[utf8]{inputenc} 
\usepackage[T1]{fontenc}    
\usepackage{hyperref}       
\usepackage{url}            
\usepackage{booktabs}       
\usepackage{amsmath}
\usepackage{graphicx}
\usepackage{amsfonts}       
\usepackage{nicefrac}       
\usepackage{microtype}      
\usepackage{amsthm}
\usepackage[justification=centering]{caption}
\newtheorem{theorem}{Theorem}
\newtheorem{lemma}{Lemma}
\newtheorem{definition}{Definition}

\newtheorem{proposition}{Proposition}

\usepackage{bbm, dsfont}

\DeclareMathOperator*{\arginf}{arg\,inf}

\begin{document}
%
\title{On the Rates of Convergence from Surrogate Risk Minimizers to the Bayes Optimal Classifier}
%
%
%

\author{Jingwei Zhang,
        Tongliang Liu,~\IEEEmembership{Member,~IEEE,}
        and~Dacheng~Tao,~\IEEEmembership{Fellow,~IEEE}
\IEEEcompsocitemizethanks{\IEEEcompsocthanksitem J. Zhang, T. Liu, and D. Tao are with the UBTECH Sydney Artiticial Intelligence Centre, and the School of Computer Science, in the Faculty of Engineering and Information Technologies, The University of Sydney, J12 Cleveland St, Darlington, NSW 2008, Australia. E-mail: zjin8228@uni.sydney.edu.au, tongliang.liu@sydney.edu.au, dacheng.tao@sydney.edu.au.}
}
%
%

\markboth{Journal of \LaTeX\ Class Files,~Vol.~14, No.~8, August~2015}%
{Shell \MakeLowercase{\textit{et al.}}: Bare Demo of IEEEtran.cls for IEEE Journals}
%



\maketitle

\begin{abstract}
In classification, the use of 0-1 loss is preferable since the minimizer of 0-1 risk leads to the Bayes optimal classifier. However, due to the non-convexity of 0-1 loss, this optimization problem is NP-hard. Therefore, many convex surrogate loss functions have been adopted. Previous works have shown that if a Bayes-risk consistent loss function is used as surrogate, the minimizer of the emprical surrogate risk can achieve the Bayes optimal classifier as sample size tends to infinity. Nevertheless, the comparison of convergence rates of minimizers of different empirical surrogate risks to the Bayes optimal classifier has rarely been studied. Which characterization of the surrogate loss determines its convergence rate to the Bayes optimal classifier? Can we modify the loss function to achieve a faster convergence rate?   In this paper, we study the convergence rates of empirical surrogate minimizers to the Bayes optimal classifier. Specifically, we introduce the notions of \emph{consistency intensity} and \emph{conductivity} to characterize a surrogate loss function and exploit this notion to obtain the rate of convergence from an empirical surrogate risk minimizer to the Bayes optimal classifier, enabling fair comparisons of the excess risks of different surrogate risk minimizers. The main result of the paper has practical implications including  (1) showing that hinge loss (SVM) is superior to logistic loss (Logistic regression) and exponential loss (Adaboost) in the sense that its empirical minimizer converges faster to the Bayes optimal classifier and (2) guiding the design of new loss functions to speedup the convergence rate to the Bayes optimal classifier with a data-dependent loss correction method inspired by our theorems.
\end{abstract}


%
\IEEEpeerreviewmaketitle

\section{Introduction}
Classification is a fundamental machine learning task used in a wide range of applications. When designing classification algorithms, the 0-1 loss function is preferred, as it helps produce the Bayes optimal classifier, which has the minimum probability of classification error. However, the 0-1 loss is difficult to optimize because it is neither convex nor smooth (\cite{BENDAVID2003496,feldman2012agnostic}). Many different computationally-friendly surrogate loss functions have therefore been proposed as approximations for the 0-1 loss function.

However, natural questions arise of whether they are good approximations, and then what the differences are between the surrogate loss functions and the 0-1 loss. To address the first question, the Bayes-risk consistency concept has been introduced (\cite{lugosi2004bayes,bartlett2006convexity}). A surrogate loss function is said to be Bayes-risk consistent if its corresponding empirical minimizer converges to the Bayes optimal classifier when the predefined hypothesis class is universal. That means, with a sufficiently large sample, the minimizers of those surrogate risks are identical to the minimizer of the 0-1 risk in the sense that they achieve the same minimum probability of classification error. Existing results (e.g., \cite{zhang2004statistical,bartlett2006convexity,pmlr-v40-Agarwal15,neykov2016characterization}) show Bayes-risk consistency under different conditions and, reassuringly, most of the frequently used surrogate loss functions are Bayes-risk consistent.

Although Bayes-risk consistency describes the interchangeable relationship between surrogate loss functions and the 0-1 loss function, it is an asymptotic concept. The non-asymptotic link between a specific surrogate loss function and the 0-1 loss function has remained elusive. In this paper, we study the rates of convergence from surrogate risk minimizers to the Bayes optimal classifier. We derive upper bounds for the difference between the probabilities of classification error w.r.t. the surrogate risk minimizer and the Bayes optimal classifier.
Specifically, we introduce the notions of \emph{consistency intensity} $I$ and \emph{conductivity} $S$, which are uniquely determined by the surrogate loss functions. We show that for any given surrogate loss function $\phi$, if the convergence rate of the excess surrogate risk $R_{\phi}(f_n)-R_{\phi}^*$ is of order $\mathcal{O}(\frac{1}{n^p})$, where $f_n$ is the empirical surrogate risk minimizer, $R_{\phi}$ is the expected surrogate risk, and $R_{\phi}^*$ is the minimal surrogate risk achievable, the corresponding convergence rate of the expected risk $R(f_n)-R^*$ is of order $\mathcal{O}(\frac{S}{n^{pI}})$, where $R$ and $R^*$ are the expected 0-1 risk and the minimal 0-1 risk achievable, respectively. The result is able to (1) describe the non-asymptotic differences between different surrogate loss functions and (2) fairly compare the rates of convergence from different empirical surrogate risk minimizers to the Bayes optimal classifier.

\begin{figure}[h]
  \includegraphics[width=0.5\textwidth]{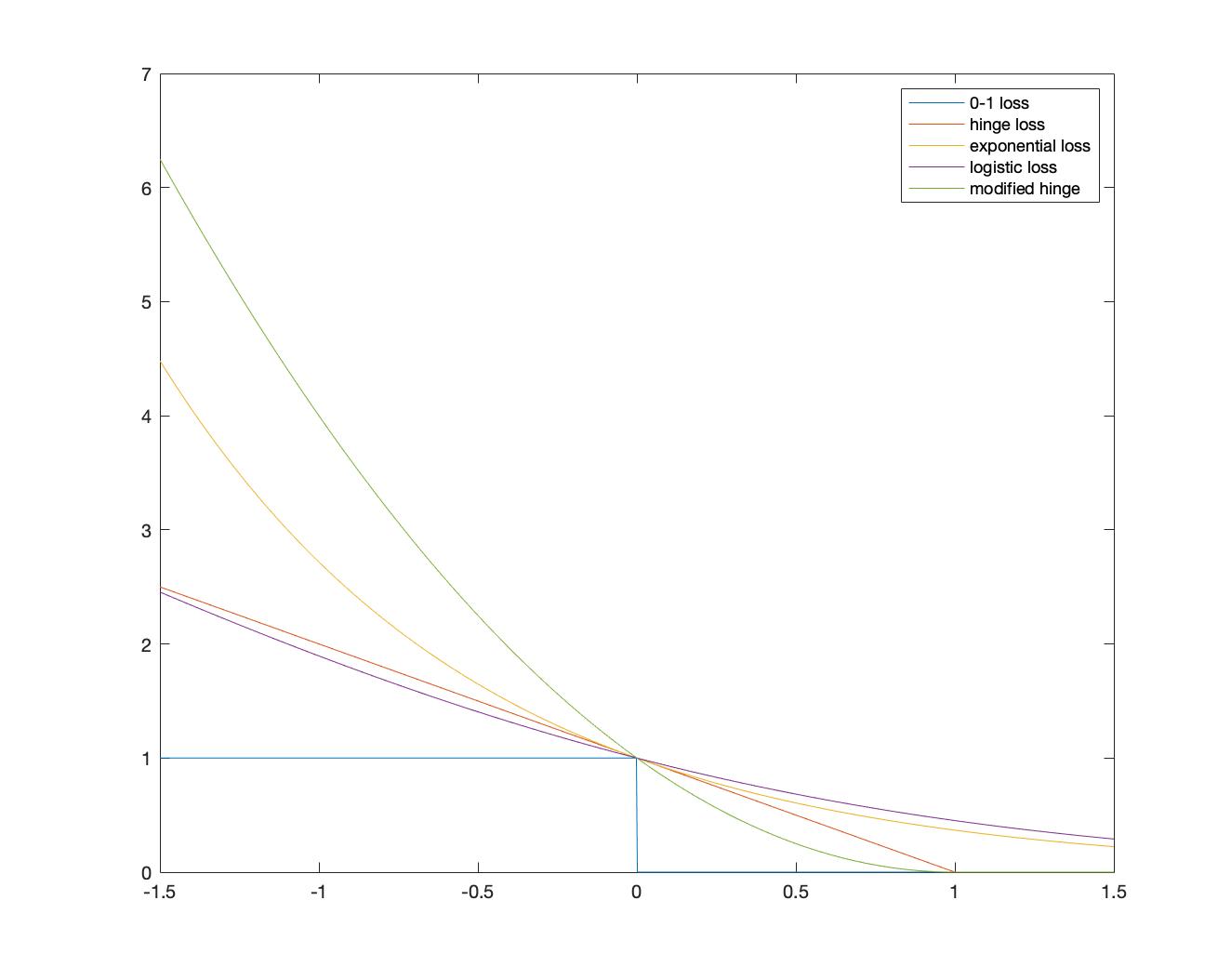}~.
  \centering
    \caption{Different convex surrogate loss functions studied in this paper. The modified hinge loss is studied in the second to last section where we take the modification parameter $\Delta=1$ in the figure.} \label{fig1}
\end{figure}

We apply our theorems to popular surrogate loss functions such as the hinge loss function in support vector machine, exponential loss function in AdaBoost, and logistic loss function in logistic regression (\cite{collins2002logistic,vapnik2013nature}), as illustrated in Figure \ref{fig1}. We conclude that SVM converges faster to the Bayes optimal classifier than AdaBoost and logistic regression, while AdaBoost and logistic regression have the same convergence rate. Furthermore, we provide a general rule for fairly comparing the convergence rates of different classification algorithms.

We show that for a data-independent surrogate loss, both the consistency intensity $I$ and conductivity $S$ are constants, and for different surrogate loss functions, $I$ and $S$ vary in $(0,1]$ and $(0, +\infty)$, respectively. Since different minimizers converge to the Bayes optimal classifier at rate $\mathcal{O}(\frac{S}{n^{pI}})$, they do not contribute to accelerating the convergence rate. However, if we modify the surrogate loss function according to the sample size $n$, $S$ can vary w.r.t. $n$. This finding enables us to devise a data-dependent loss modification method which can accelerate the convergence of a surrogate risk minimizer to the Bayes optimal classifier.

\textbf{Advantages and Disadvantages.}  Despite extensive existing works that study the convergence of surrogate risk minimizer to the Bayes optimal classifier, they are focusing on the convergence of some specific loss functions satisfying certain regularity conditions, such as convex and differentiable. Besides, these works do not show which characterization of the given surrogate loss function determines the convergence rate. Furthermore, existing work does not show how to compare the convergence rate to the Bayes optimal classifier when given several different surrogate loss functions. Another advantage of our result is that we can modify the loss function to achieve a faster convergence rate to the Bayes optimal classifier, and the modification of loss is data-dependent. Hence, our theorem provides a novel way to design better surrogate loss functions such that the minimizer converges faster to the Bayes optimal classifier.  In spite of such advantages in our work, there are also some critical problem that remains unsolved: what is the fastest convergence rate that is achievable by any surrogate loss function?  To answer this problem, we need to derive a lower bound of the convergence rate under certain regularity conditions on the loss function, which leaves as future work.
 
\textbf{Organization.} The remainder of the paper is organized as follows. In section \ref{setup}, we first introduce basic mathematical notations. Then we introduce the notions of Bayes optimality and Bayes-risk consistency, along with several lemmas and related works necessary to the proofs of our main theorem. We present our main theorem in section \ref{sec3}, namely the rate of convergence from the expected risk to the Bayes risk for empirical surrogate risk minimization. In section \ref{sec4}, we show several applications of our theorem. In particular, section \ref{app1} details several specific examples of computing and comparing the rates of convergence to the Bayes risk and provides a general rule to compare the rates of convergence to the Bayes risk for any two algorithms with Bayes-risk consistent loss functions. Then, in section \ref{app2}, we apply our theorems to modifying the hinge loss function to accelerate its convergence to the Bayes risk. In section \ref{sec5}, we conclude the paper and briefly discuss future works.

\section{Preliminaries} \label{setup}
We present basic notations in Section \ref{notation} and briefly introduce the concept of Bayes-risk consistency and necessary lemmas in Section \ref{bayes}. In Section \ref{priorwork}, we discuss related works about the statistical properties of Bayes-risk consistent surrogate loss functions, which play critical roles in the proof of our theorems.
\subsection{Notation}\label{notation}
In this paper, we focus on binary classification, where the feature space $\mathcal{X}$ is a subset of a Hilbert Space $\mathcal{H}$ and the label space is denoted by $\mathcal{Y}=\{-1,+1\}$. We assume that a pair of random variables $(X,Y)$ is generated according to an unknown distribution $D$, where $P(X,Y)$ is the corresponding joint probability. Binary classification aims to find a map $f: \mathcal{X}\rightarrow \mathbb{R} $ within some particular predefined hypothesis class $\mathcal{F}$ such that the sign of $f(X)$ can be used as a prediction for $Y\in\mathcal{Y}$.

To evaluate the goodness of $f$, some performance measures are required. Intuitively,
the 0-1 risk is employed, defined as:
\begin{equation}
R(f) = \mathbb{E}[\mathbbm{1}[sgn(f(X))\neq Y]]=P(sgn(f(X))\neq Y)
\end{equation}
where $\mathbbm{1}[\cdot]$ denotes the indicator function. From the definition, we can see that minimizing the 0-1 risk is equivalent to minimizing the probability of classification error.  We hope to find the function such that the probability of classification error is minimized, which is called the \emph{Bayes optimal classifier}, defined as follows:
\begin{equation}
f^{*}=\arginf_{f} R(f)
\end{equation}
where the infimum is over all measurable functions. The corresponding expected risk is called the Bayes risk:
\begin{equation}
R^{*}= R(f^{*})~.
\end{equation}
In this paper, we assume that the predefined hypothesis class is universal, which means the Bayes optimal classifier is always in $\mathcal{F}$.

Since the joint distribution $D$ is unknown, we cannot calculate $R(f)$ directly. Given a training sample $\bold{S}_n=\{(x_1,y_1),...,(x_n,y_n)\}$, the following empirical risk is widely exploited to approximate the expected risk $R$:
\begin{equation}
\hat{R}(f)=\frac{1}{n} \sum_{i=1}^{n} \mathbbm{1}[sgn(f(x_i))\neq y_i]~.
\end{equation}
Directly minimizing the above empirical risk is NP-hard due to the non-convexity of the 0-1 loss function, which forces us to adopt convex surrogate loss functions. Similarly, for any non-negative surrogate loss function $\phi: \tilde{\mathcal{Y}}\times\mathcal{Y}\rightarrow\mathbb{R}^{+}$, where $\tilde{\mathcal{Y}}$ is the space of the output of the classifier, we can define the $\phi$-risk, optimal $\phi$-risk, empirical $\phi$-risk, and the empirical surrogate risk minimizer as:
\begin{eqnarray}
&&R_\phi(f) = \mathbb{E}[\phi(f(X),Y))]~,\\
&&R_\phi^{*} = \inf_{f} R_\phi(f)~,\\
&&\hat{R}_{\phi}(f)=\frac{1}{n} \sum_{i=1}^{n} \phi(f(x_i),y_i)~,\\
&&f_n=\arginf_{f\in\mathcal{F}}\hat{R}_{\phi}(f)~.
\end{eqnarray}
Moreover, we define the \emph{excess risk} and the \emph{excess $\phi$-risk}, respectively, as follows:
\begin{eqnarray}
&&R(f_n)-R^*,\\ 
&&R_{\phi}(f_n)-R_{\phi}^{*}~.
\end{eqnarray}
For classification tasks, the loss function is often margin-based and we can rewrite  $\phi(f(x_i),y_i)$ as $\phi(y_if(x_i))$, where the quantity $yf(x)$ is known as the margin, which can be interpreted as the confidence in prediction (\cite{mohri2012foundations}). In this paper, we study margin-based loss functions.

\subsection{Optimality and Bayes-risk Consistency} \label{bayes}
We first introduce the concept of \emph{Bayes optimal}. Let define
\begin{equation}
\eta(X)=P(Y=1|X)~.
\end{equation}
\begin{lemma} (\cite{bousquet2004introduction}) \label{lemma1}
Assume the random pair $(X,Y)\in\mathcal{X}\times\mathcal{Y}$ follows a given distribution $D$. Then, any classifier $f:\mathcal{X}\rightarrow\mathbb{R}$, which satisfies $sgn(f(X))=sgn(\eta(X)-1/2)$, is Bayes optimal under D.
\end{lemma}
Before we introduce the notion of Bayes-risk consistency for surrogate loss functions, we need to present several basic definitions (\cite{bousquet2004introduction}). First, we define the \emph{ conditional $\phi$-risk }as:
\begin{equation}
\mathbb{E}[\phi(Yf(X))|X=x]=\eta(x) \phi(f(x))+(1-\eta(x))\phi(-f(x))~.
\end{equation}
We then introduce the \emph{ generic conditional $\phi$-risk} by letting $z=f(x)$ and $\eta=\eta(x)$:
\begin{equation}
C(\eta,z)=\eta \phi(z)+(1-\eta)\phi(-z)~.
\end{equation}
Note that the function $C(\eta,z)$ is a convex combination of $\phi(z)$ and $\phi(-z)$. It immediately follows the definition of \emph{optimal generic conditional $\phi$-risk }:
\begin{equation}
C^{*}(\eta)=\inf_{z\in\mathbb{R}}\eta \phi(z)+(1-\eta)\phi(-z)~.
\end{equation}
We also define $C^{*-}(\eta)$ as:
\begin{equation}
C^{*-}(\eta)=\inf_{z:z(\eta -1/2)\leq 0}C_{\eta}(z)~.
\end{equation}
This definition follows the \emph{optimal generic conditional $\phi$-risk }, but with the constraint that the sign of the output $z$ differs from $sgn(\eta-1/2)$.  Under these settings, we define the Bayes-risk consistency.
\begin{definition} (\cite{lugosi2004bayes})
A surrogate loss function $\phi$ is Bayes-risk consistent if the minimizer of the conditional $\phi$-risk, $f^*=\arginf_{f}\mathbb{E}[\phi(Yf(X))|X=x]$, has the same sign as the Bayes optimal classifier for any $x\in \mathcal{X}$. Or simply, $sgn(f^*(x))=sgn(\eta(x)-1/2)$.
\end{definition}
We here present a necessary and sufficient condition for the Bayes-risk consistency of surrogate loss functions.
\begin{lemma} (\cite{bartlett2006convexity})\label{pp3}
A surrogate loss is Bayes-risk consistent if and only if $C^{*}(\eta)< C^{*-}(\eta)$, for any $\eta\neq 1/2$.
\end{lemma}
Lemma \ref{pp3} has an intuitive explanation: any Bayes-risk consistent loss requires the constraint that it always leads to strictly larger conditional $\phi$-risks for any $X$ when the signs of the output $f(X)$ differs from that of Bayes optimal classifier.

\subsection{Asymptotic Consistency in Surrogate Risk Optimization} \label{priorwork}
We briefly introduce some related works on Bayes-risk consistency and its statistical properties in classification. \cite{lugosi2004bayes} proved that the Bayes-risk consistency is satisfied for empirical surrogate loss minimization under the condition that the surrogate loss $\phi$ is strictly convex, differentiable, and monotonic with $\phi(0)=1$. \cite{bartlett2006convexity} then offered a more general result, showing that the Bayes-risk consistency is possible if and only if the loss function is Bayes-risk consistent\footnote{In their paper, the term ``classification-calibrated" is used instead of ``Bayes-risk consistent".}. Other results on Bayes-risk consistency under different assumptions have been presented by \cite{zhang2004statistical,steinwart2005consistency,neykov2016characterization}.

Below, we give the main results proved in \cite{bartlett2006convexity}, which shows that minimizing over any surrogate risk with Bayes-risk consistent loss function is asymptotically equivalent to minimizing the 0-1 risk and thus leads to the Bayes optimal classifier.
\begin{theorem} (\cite{bartlett2006convexity})\label{t1}
For any convex loss function $\phi$, it is Bayes-risk consistent if and only if it is differentiable at $0$ and $\phi'(0)<0$. Then for such a convex and Bayes-risk consistent loss function $\phi$, any measurable function $f:\mathcal{X}\rightarrow\mathbb{R}$, and any distribution $P$ over $\mathcal{X}\times\mathcal{Y}$, the following inequality holds,
\begin{equation}\label{eq0}
\psi(R(f)-R^*)\leq R_{\phi}(f)-R_{\phi}^*
\end{equation}
where $\psi(\theta)=\phi(0)-C^{*}(\frac{1+\theta}{2})$ is nonnegative, convex and invertible on $[0,1]$ and has only one zero at $\theta=0$.
\end{theorem}
The above theorem gives an upper bound on the excess risk in terms of the excess $\phi$-risk and shows that minimizing over any convex Bayes-risk consistent surrogate loss is asymptotically equivalent to minimizing over 0-1 loss because the function $\psi$ is invertible and only have a single zero at $\theta = 0$. We provide deeper insights of this theorem in the next section.

\section{The Rates of Convergence from Empirical Surrogate Risk Minimizers to the Bayes Optimal Classifier} \label{sec3}
We know that optimizing over any empirical surrogate risk with a Bayes-risk consistent loss function will lead to the Bayes optimal classifier when the training sample size $n$ is large enough. However, a natural question arises when we optimize over different empirical surrogate risks with Bayes-risk consistent loss functions: ``What are the difference between them? Do the minimizers have the same rate of convergence to the Bayes optimal classifier?" Those problems are essential because when we choose classification algorithms for a real-world problem, we may expect a fast convergence to the Bayes optimal classifier which also implies a small sample complexity.

To answer the above questions, we need to find a proper metric to measure the distance between $f_n$ and  $f^{*}$. Since we are mainly care about the probability of classification error of the proposed learning algorithm, it is reasonable to measure the rate of convergence from the expected risk $R(f_n)$ to the Bayes risk $R^*$ instead of measuring the rate of convergence from $f_n$ to $f^{*}$ directly.

Before showing the results of convergence rates, we present the intuition of our work.
\subsection{Intuition of the Proposed Method}
For any empirical surrogate risk minimizer $f_n$, we can rewrite the inequality in Theorem \ref{t1} as follows:
\begin{equation}
\begin{aligned}
&\psi(R(f_n)-R^*)\leq R_{\phi}(f_n)-R_{\phi}^*\\&=\left(R_{\phi}(f_n)-\inf_{f\in\mathcal{F}}R_{\phi}(f)\right)+\left(\inf_{f\in\mathcal{F}}R_{\phi}(f)-R_{\phi}^*\right) \label{eq1}~.
\end{aligned}
\end{equation}
To achieve our goal of measuring the rate of convergence from  the empirical surrogate risk minimizer to the Bayes optimal classifier, we need to bound the term $R(f_n)-R^*$. As $\psi(\theta)$ is invertible, we can bound the term on the right hand side of the inequality. We call the first term  in the right hand side the \emph{estimation error}, which depends on the learning algorithm and the training data. The second term is called the \emph{approximation error}, depending on the choice of the hypothesis class $\mathcal{F}$. Often, the hypothesis class is predefined and universal. Thus, in this paper, we just assume that the Bayes optimal classifier is right in the hypothesis class $\mathcal{F}$. In other words, we have $\inf_{f\in\mathcal{F}}R_{\phi}(f)=R_{\phi}^*$ and (\ref{eq1}) becomes
\begin{equation}
\psi(R(f_n)-R^*)\leq R_{\phi}(f_n)-\inf_{f\in\mathcal{F}}R_{\phi}(f)~.
\end{equation}
The right side of the above inequality can be further upper bounded by
\begin{equation}
R_{\phi}(f_n)-\inf_{f\in\mathcal{F}}R_{\phi}(f) \leq 2\sup_{f\in\mathcal{F}}|\hat{R}_{\phi}(f)-R_{\phi}(f)|~,
\end{equation}
where the defect on the right hand side is called the \emph{generalization error}.
Using the concentration of measure (\cite{boucheron2013concentration}), and the uniform convergence argument, e.g., VC-dimension (\cite{vapnik2013nature}), covering number (\cite{zhang2002covering}), and Rademacher complexity (\cite{bartlett2002rademacher}), the generalization error can be non-asymptotically upper bounded with a high probability. Often, the upper bounds can reach the order $\mathcal{O}(\frac{1}{\sqrt{n}})$ (\cite{mohri2012foundations}).
We also notice that the excess $\phi$-risk can achieve convergence rates faster than $\mathcal{O}(\frac{1}{\sqrt{n}})$, such as exploiting local Rademacher complexities, low noise models, and strong convexity \cite{tsybakov2004optimal,bartlett2005local,koltchinskii2006local,NIPS2008_3400,liu2017algorithmic}. Here we mainly consider the ordinary case where the convergence rate of the excess $\phi$-risk is of order $\mathcal{O}(\frac{1}{\sqrt{n}})$, but our results can directly generalize to other cases.

Theorem \ref{t1} also implies that if $\phi$ is convex and Bayes-risk consistent, then for any sequence of measurable functions $f_i: \mathcal{X}\rightarrow\mathbb{R}$ and any distribution $P$ over $\mathcal{X}\times\mathcal{Y}$,
\begin{equation} \label{eq2}
R_{\phi}{(f_i)}\rightarrow R_{\phi}^{*}\quad \Rightarrow \quad R(f_i)\rightarrow R^{*}~.
\end{equation}
This presents the dynamics of the Bayes risk consistency for any convex Bayes-risk consistent loss in an asymptotic way.

Observe that in equation (\ref{eq0}), the asymptotic consistency property is mainly due to the uniqueness of the zero of the function $\psi(\theta)$, where the only zero of function $\psi(\theta)$ is at $\theta = 0$. Thus, when $ R(f_i) - R^{*} \rightarrow 0$, we have $\psi(R{(f_i)}- R^{*})\rightarrow 0$, which leads to $R{(f_i)}\rightarrow R^{*}$. However, if we want to derive the rate of convergence from the expected risk $R{(f_n)}$ to the Bayes risk $R^{*}$, we may need more detailed or higher order properties of the function $\psi(\theta)$ in the infinitesimal right neighborhood of $\theta = 0$, rather than just the value of $\psi(\theta)$ at $\theta=0$.

In the next subsection, we will exploit the upper bound of $R_{\phi}(f_n)-R_{\phi}^*$ and a higher order property of the function $\psi(\theta)$ to derive upper bounds for $R{(f_n)}- R^{*}$.

\subsection{Consistency Intensity for Bayes-risk Consistent Loss Functions}
Knowing the convergence rate of the excess $\phi$-risk and their relation $\psi(R(f_n)-R^*)\leq R_{\phi}(f_n)-R_{\phi}^*$, it's straightforward to consider taking an inverse of the function $\psi$ on the both sides, yielding an upper bound, $\psi^{-1}(R_{\phi}(f)-R_{\phi}^*)$. However, in reality, for most of Bayes-risk consistent loss functions, the corresponding $\psi(\theta)$ is sometimes intractable to take an inverse analytically. Furthermore, the term $\psi^{-1}(\mathcal{O}(\frac{1}{\sqrt{n}}))$ may not reflect the order of $n$ explicitly. Thus, we must figure out the factor that determines the convergence rate of the excess risk $R(f_n)-R^*$---that is some high order property of function $\psi(\theta)$ within the infinitesimal right neighborhood of $\theta=0$. Before we move on to our main theorems, let's introduce some basic lemmas and propositions first. 
\begin{proposition}\label{pp1}
For any two functions $f(\theta)$ and $g(\theta)$, which are differentiable at $\theta=0$ and satisfy $f(0)=g(0)=0$, then, the following conditions are equivalent:
\begin{itemize}
\item $\lim_{\theta\rightarrow 0}\frac{f(\theta)}{g(\theta)}=A$~;
\item $f(\theta)=Ag(\theta)+o(g(\theta))$~;
\item $f(\theta)=\mathcal{O}(g(\theta))$ when $A \neq 0$~.
\end{itemize}
\end{proposition}
Proposition \ref{pp1} can be proved directly by following the definition of the $o$ and $\mathcal{O}$ notation. We now introduce the notions of consistency intensity and conductivity for convex Bayes-risk consistent loss functions.
\begin{lemma}\label{df2}
For any given convex  Bayes-risk consistent loss function $\phi$, let $\psi(\theta)=\phi(0)-C^{*}(\frac{1+\theta}{2})$. There exists two unique constant $\alpha\in\mathbb{R}^{+}$ and $M\in\mathbb{R}^{+}$ such that
\begin{equation}\label{eq7}
\lim_{\theta\rightarrow 0+}\frac{\psi(\theta)}{M\theta^{\alpha}}=1~.
 \end{equation}
 We call $I=\frac{1}{\alpha}$ the consistency intensity of this Bayes-risk consistent loss function and $S=M^{-\frac{1}{\alpha}}$ the conductivity of the intensity.
 \begin{proof}
 From Theorem 1,  we know that $\psi(\theta)$ is a convex function. It's also known that any convex function on a convex open subset of $\mathbb{R}^n$ is semi-differentiable. Thus, we can  denote the right derivative of $\psi(\theta)$ at $\theta=0$ by $\partial_{+}\psi(0)$. Using Maclaurin expansion, we have,
 \begin{equation}
 \psi(\theta)=\psi(0)+\partial_{+}\psi(0)*\theta+o(\theta)=\partial_{+}\psi(0)*\theta+o(\theta)~.
 \end{equation}
Followed by Proposition \ref{pp1}, 
if $\partial_{+}\psi(0)\neq 0$, we have,
\begin{equation}
 M=\partial_{+}\psi(0) \quad \text{and}\quad \alpha=1~.
 \end{equation}
 If $\partial_{+}\psi(0)= 0$, then $\psi(\theta)=o(\theta)$, which means that $\psi(\theta)$ is the infinitesimal of higher order than $\theta$ as $\theta\rightarrow 0+$. Then, by definition, for any given $\phi$, we can compute $\psi$. Because $\psi(\theta)$ is the higher order infinitesimal of $\theta$ as $\theta\rightarrow 0+$, there exist unique $\alpha>1 $ and $M\in\mathbb{R}^{+}$ such that,
 \begin{equation}
\lim_{\theta\rightarrow 0+}\frac{\psi(\theta)}{M\theta^{\alpha}}=1
 \end{equation}
 which completes the proof.
 \end{proof}
\end{lemma}
We then introduce Lemmas \ref{lm8} and \ref{lm3}, which are essential to the proof of our main theorems.
\begin{lemma}\label{lm8}
Given any convex Bayes-risk consistent loss function $\phi$, let the inverse of its corresponding $\psi$-transform be denoted by $\psi^{-1}$. We have
\begin{equation}\label{lm1}
\lim_{\mu\rightarrow 0+}\frac{\psi^{-1}(\mu)}{S\mu^{I}}=1~.
\end{equation}
\begin{proof}
Let $\psi^{-1}(\mu)=\theta$, then $\mu=\psi(\theta)$.
 From Theorem \ref{t1}, we know that $\psi$ is monotonic increasing within $[0,1]$, and $\psi(0)=0$. Thus,
 \begin{center}
 $\mu\rightarrow 0+$
\quad  implies \quad
   $\theta\rightarrow 0+~.$
   \end{center}
We have,

\begin{equation}
\lim_{\mu\rightarrow 0+}\frac{\psi^{-1}(\mu)}{S\mu^{I}} =\lim_{\theta\rightarrow 0+}\frac{\theta}{S(\psi(\theta))^{I}} \label{eq6}~.
\end{equation}
By substituting the definitions of $S$ and $I$, the right hand side of (\ref{eq6}) becomes,
\begin{equation}
\lim_{\theta\rightarrow 0+}\frac{M^{\frac{1}{\alpha}}\theta}{(\psi(\theta))^{\frac{1}{\alpha}}}
=\left[\lim_{\theta\rightarrow 0+}\frac{M\theta^{\alpha}}{\psi(\theta)}\right]^{\frac{1}{\alpha}}=1~.
\end{equation}
The last equality follows (\ref{eq7}) in Lemma \ref{df2}, which completes the proof.
\end{proof}
\end{lemma}

 Lemma \ref{lm8} shows that the equivalent infinitesimal of $\psi^{-1}$ near $0$ is $S\mu^{I}$. Then, we introduce an important property for the $\psi$-transform.
 \begin{lemma}\label{lm3}
Given any convex Bayes-risk consistent loss function $\phi$, its corresponding function $\psi^{-1}$ is interchangeable with $\mathcal{O}(\frac{1}{n^p})$ for any $p>0$. That is,
\begin{equation}\label{lm2}
\psi^{-1}\left(\mathcal{O}(\frac{1}{n^p})\right)=\mathcal{O}\left(\psi^{-1}(\frac{1}{n^p})\right)~.
\end{equation}
\begin{proof}
From Proposition \ref{pp1}, we have that there exists $0<A,B<+\infty$ such that,
\begin{equation}
\mathcal{O}\left(\frac{1}{n^p}\right)=A\frac{1}{n^p}+o\left(\frac{1}{n^p}\right)\label{eq10}
\end{equation}
and
\begin{equation}
\mathcal{O}\left(\psi^{-1}(\frac{1}{n^p})\right)=B\psi^{-1}\left(\frac{1}{n^p}\right)+o\left(\psi^{-1}(\frac{1}{n^p})\label{eq11}\right)
~.\end{equation}
Substituting (\ref{eq10}) and (\ref{eq11}) into ({\ref{lm2}}), it's equivalent to proving that  for any $0<A<+\infty$, there exists $0<B<+\infty$, such that,
\begin{equation}
\psi^{-1}\left(A\frac{1}{n^p}+o(\frac{1}{n^p})\right)=B\psi^{-1}\left(\frac{1}{n^p}\right)+o\left(\psi^{-1}(\frac{1}{n^p})\label{eq12}\right)~.
\end{equation}
 To prove (\ref{eq12}), by Proposition \ref{pp1}, we only need to prove that, for any $0<A<+\infty$, there exists $0<B<+\infty$, such that,
\begin{equation}\label{eq13}
\lim_{n\rightarrow+\infty} \frac{\psi^{-1}(A\frac{1}{n^p}+o(\frac{1}{n^p}))}{\psi^{-1}(\frac{1}{n^p})}=B~.
\end{equation}
Followed by Lemma \ref{lm8} and proposition \ref{pp1}, we have
\begin{equation}
\psi^{-1}(\mu)=S\mu^{I}+o(\mu^I)~.   \label{eq9}
\end{equation}
Substituting (\ref{eq9}) into (\ref{eq13}), we have,
\begin{equation}
\begin{aligned}
&\lim_{n\rightarrow+\infty} \frac{\psi^{-1}(A\frac{1}{n^p}+o(\frac{1}{n^p}))}{\psi^{-1}(\frac{1}{n^p})}
\\&=\lim_{n\rightarrow+\infty} \frac{S(A\frac{1}{n^{p}}+o(\frac{1}{n^{p}}))^I+o((A\frac{1}{n^{p}}+o(\frac{1}{n^{p}}))^I)}{S(\frac{1}{n^{p*I}})+o((\frac{1}{n^{p*I}}))}
\\&=\lim_{n\rightarrow+\infty} \frac{S(A^{I}\frac{1}{n^{p*I}}+o(\frac{1}{n^{p*I}}))}{S(\frac{1}{n^{p*I}})}
\\&=A^{I}~.
\end{aligned}
\end{equation}
This means that for any $0<A<+\infty$, there exists $B=A^I\in (0, +\infty)$ such that (\ref{eq13}) holds true, which completes the proof.
\end{proof}
\end{lemma}

\noindent Following the definitions of $S$ and $I$ in Lemma \ref{df2}, we have our first main theorem.
\begin{theorem}\label{thm2}
Suppose the excess $\phi$-risk satisfies $R_{\phi}(f_n)-R_{\phi}^*\leq\mathcal{O}(\frac{1}{n^p})$ with a high probability. Then, with the same high probability, we have
\begin{equation}
 R(f_n)-R^*\leq\mathcal{O}\left(\frac{S}{n^{pI}}\right)~.
 \end{equation}
 \begin{proof}
 From Theorem \ref{t1}, we have that
 \begin{equation}
  R(f_n)-R^*\leq \psi^{-1}(R_{\phi}(f_n)-R_{\phi}^*)
 \end{equation}
 where $f_n$ is the minimizer of the empirical surrogate risk $\hat{R}_{\phi}$ with sample size $n$. Under our assumption that the Bayes optimal classifier is within the hypothesis class $\mathcal{F}$, then, with a high probability, we have (\cite{bousquet2004introduction}),
  \begin{equation}
R_{\phi}(f_n)-R_{\phi}^* \leq \mathcal{O}\left(\frac{1}{n^p}\right)~. \label{eq17}
  \end{equation}
 Note that p is often equal to $1/2$ for the worst cases. With (\ref{eq17}) and Lemma \ref{lm3}, we have,
    \begin{equation}\label{eeq23}
  \begin{aligned}
&\psi^{-1}(R_{\phi}(f_n)-R_{\phi}^*) \\& \leq \psi^{-1}\left(\mathcal{O}(\frac{1}{n^p})\right)=\mathcal{O}\left(\psi^{-1}(\frac{1}{n^p})\right)~.
\end{aligned}
  \end{equation}
 Substituting (\ref{eq9}) into (\ref{eeq23}), we have,
 \begin{equation}
 \begin{aligned}
\\& \mathcal{O}\left(\psi^{-1}(\frac{1}{n^p})\right)=\mathcal{O}\left(S(\frac{1}{n^p})^I+o((\frac{1}{n^p})^I)\right)\\&=\mathcal{O}\left(\frac{S}{n^{pI}}\right)~,
 \end{aligned}
\end{equation}
which completes the proof.
 \end{proof}
\end{theorem}

From Theorem \ref{thm2}, we show that the consistency intensity $I$ and conductivity $S$ have direct influence on the convergence rate, which is of order $\mathcal{O}(\frac{S}{n^{pI}})$. When $0<I<1$, the convergence rate from $R(f_n)$ to $R^*$ will be slower than the convergence rate from $R_{\phi}(f_n)$ to $R_{\phi}^*$; if $I=0$, the algorithm will never reach the Bayes optimal classifier; if $I=1$, the convergence rates will be the same. In the next theorem, we will show that for any convex Bayes-risk consistent loss function, the range of $I$ is $(0,1]$.
\begin{theorem} \label{thm3}
For any convex Bayes-risk consistent loss function $\phi$,  it always holds true that $0<I\leq 1$.

\begin{proof}
We have that $0<\alpha<+\infty$, so $I>0$ holds trivially. To prove $I=\frac{1}{\alpha}\leq1$, it is equivalent to proving that there exists $0\leq C<+\infty$ such that,
\begin{equation}
\psi(\theta)=C\theta+o(\theta) \label{eq20}~,
\end{equation}
because $I<1$ holds true if and only if $C=0$; and $I=1$ holds true if and only if $0<C<+\infty$. From proposition \ref{pp1}, the equation (\ref{eq20}) implies,
\begin{equation}
\lim_{\theta\rightarrow 0+}\frac{\psi(\theta)}{\theta}=C~.
\end{equation}
Since any convex function on a convex open subset in $\mathbb{R}^n$ is semi-differentiable, $\psi(\theta)$ is at least right differentiable at $\theta=0$. We therefore have,
 \begin{equation}
\lim_{\theta\rightarrow 0+}\frac{\psi(\theta)}{\theta}=\lim_{\theta\rightarrow 0+}\frac{\psi(\theta)-\psi(0)}{\theta-0}
=\partial_{+}\psi(0)=C
\end{equation}
where $\partial_{+}\psi(0)$ denotes the right derivative of $\psi(\theta)$ at $\theta=0$. The proof ends.
\end{proof}
\end{theorem}
As will be shown in the later examples, for Adaboost (exponential loss) and Logistic regression (logistic loss), we have $C=0$, then $I$ will be strict less than one. For SVM (Hinge Loss), we have $C>0$, then by definition we have $I=1$.  Theorem \ref{thm3} shows that for data-independent surrogate loss functions, because $I\leq1$, the convergence rate from $R(f_n)$ to $R^*$ will not be faster than the convergence rate from $R_{\phi}(f_n)$ to $R_{\phi}^*$. As $I>0$, it also means that optimizing over any convex Bayes-risk consistent surrogate loss will finally make the excess risk  $R(f_n)-R^*$  converge to $0$ and thus the output is the Bayes optimal classifier as sample size $n$ tends to infinity. This result matches our common sense: while we benefit from the computational efficiency of convex surrogate loss functions, we also suffer from a slower rate of convergence to the Bayes optimal classifier.

\section{Applications} \label{sec4}
In this section, we present several applications of our results. In Section \ref{app1}, we use the notion of consistency intensity to measure the rates of convergence from the empirical surrogate risk minimizers to the Bayes optimal classifier for different classification algorithms, such as support vector machine, boosting, and logistic regression. We also derive a general discriminant rule for comparing the convergence rates for different leaning algorithms.
In Section \ref{app2}, we show that the notions of consistency intensity and conductivity can help to modify surrogate loss functions so as to achieve a faster convergence rate from $R(f_n)$ to $R^*$.

\subsection{Consistency Measurement}\label{app1}
In this subsection, we first apply our results to some popular classification algorithms.

\begin{figure}[h] 
  \includegraphics[width=0.5\textwidth]{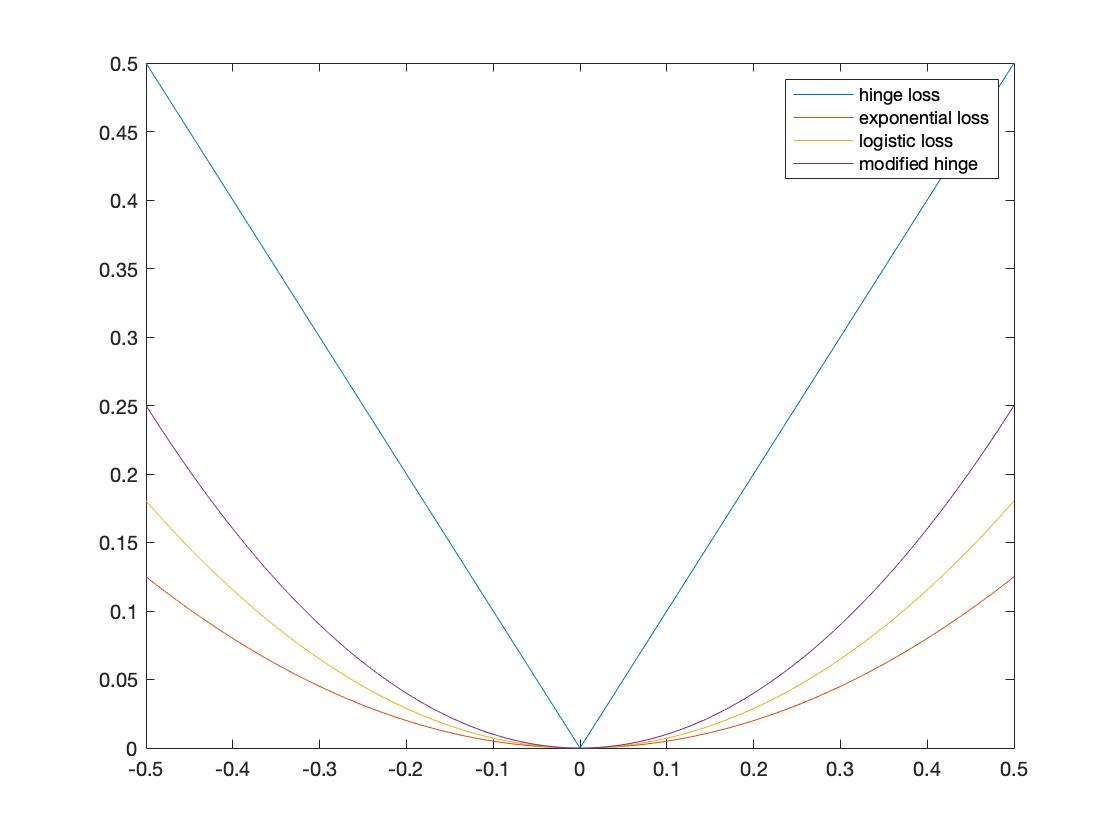}~.
  \centering
    \caption{The leading term of $\psi(\theta)$ for different loss functions, where we take $\Delta=1$ for the modified hinge loss.}\label{fig2}
\end{figure}

\emph{Example 1} (Hinge loss in SVM). Here we have $\phi(z)=max\{0,1-z\}$, which is convex and $\phi'(0)=-1<0$. Note that we have defined $C^{*}(\eta)=\inf_{z\in\mathbb{R}}\eta \phi(z)+(1-\eta)\phi(-z)$. Let
\begin{equation}
z^{*}(\eta)=\arginf_{z\in\mathbb{R}}\eta \phi(z)+(1-\eta)\phi(-z)~.
\end{equation}
It is easy to verify that
\begin{equation} 
z^{*}(\eta)= sgn(\eta-1/2)
\end{equation}
and
\begin{equation}
C^{*}(\eta)=\min\{2\eta,2(1-\eta)\}~.
\end{equation}
So we have,
\begin{equation}
\psi(\theta)=|\theta|~.
\end{equation}
Followed by theorem \ref{thm2}, we have $S=1$ and $I=1$. Thus for SVM, with a high probability, we have,
\begin{equation}
 R(f_n)-R^{*}\leq \mathcal{O}\left(\frac{1}{n^p}\right)~.
\end{equation}
\noindent
\emph{Example 2} (Exponential loss in Adaboost). We have $\phi(z)=e^{-z}$, which is convex and $\phi'(0)=-1<0$. Then, it's easy to derive that
\begin{equation}
z^{*}(\eta)= \frac{1}{2}\ln(\frac{\eta}{1-\eta})
\end{equation}
and
\begin{equation}
C^{*}(\eta)=2\sqrt{\eta(1-\eta)}~.
\end{equation}
So,
\begin{equation}
\psi(\theta)=1-\sqrt{1-\theta^2}~.
\end{equation}
Using Maclaurin expansion, we have,
\begin{equation}
\psi(\theta)=\frac{1}{2}\theta^2+o(\theta^2)~.
\end{equation}
Thus, $S=\sqrt{2}$ and $I=\frac{1}{2}$.  For Adaboost, with a high probability, we have
\begin{equation}
 R(f_n)-R^{*}\leq \mathcal{O}\left(\frac{\sqrt{2}}{n^{\frac{p}{2}}}\right)~.
\end{equation}
\noindent
\emph{Example 3} (Logistic loss in Logistic Regression). We have $\phi(z)=\log_2(1+e^{-z})$, which is convex and  $\phi'(0)=-\frac{1}{2\ln2}<0$, we can follow similar procedures as before,
\begin{equation}
C^{*}(\eta)=-\eta \log_2\eta-(1-\eta)\log_2(1-\eta)~.
\end{equation}
So,
\begin{equation}
\psi(\theta)=1-\frac{1+\theta}{2}\log_2{\frac{2}{1+\theta}}-\frac{1-\theta}{2}\log_2{\frac{2}{1-\theta}}~.
\end{equation}
Using Maclaurin expansion, we have,
\begin{equation}
\psi(\theta)=\frac{1}{2\ln2}\theta^2+o(\theta^2)~.
\end{equation}
Therefore $S=\sqrt{2\ln2}$ and $I=\frac{1}{2}$.  For Logistic Regression, with a high probability, we have
\begin{equation}
 R(f_n)-R^{*}\leq \mathcal{O}\left(\frac{\sqrt{2\ln2}}{n^{\frac{p}{2}}}\right)~.
\end{equation}
From the above examples, we conclude that SVM has a faster convergence rate to the Bayes optimal classifier than Adaboost and Logistic Regression. For better illustrations, we draw the leading term of the expansion of $\psi(\theta)$ around zero for these loss functions, as shown in Figure \ref{fig2}. It is obvious that the convergence behavior of the loss function is determined by the leading term of the expansion of $\psi(\theta)$ around zero. We note that for hinge loss, the leading term is sharper than other loss, which leads to a faster convergence, while other loss have similar asymptotic behavior around zero, which leads to the same but slower convergence rate to the Bayes optimal classifier. 

Note that the proposed methods also apply to many other surrogate loss functions.
However, if we only need to compare the convergence rates of any two classification algorithms, we may not need to compute the consistency intensity of the two surrogate loss functions explicitly. The following theorem gives a general discriminant rule.

\begin{theorem} \label{thm4}
Given two convex Bayes-risk consistent loss functions $\phi_1$ and $\phi_2$, denoting their corresponding $\psi$-transform by $\psi_1$ and $\psi_2$, we assume that their excess $\phi$-risk have the same convergence rate of order $\mathcal{O}(\frac{1}{n^p})$. Then, we define the intensity ratio $\lambda$ as follows,
\begin{equation}
\lambda=\lim_{\theta\rightarrow 0+}{\frac{\psi_1(\theta)}{\psi_2(\theta)}}~.
\end{equation}
We have the following statements:
\begin{itemize}
\item if $0<\lambda<+\infty$, then the minimizers w.r.t. $\phi_1$ and $\phi_2$ converge equally fast to the Bayes optimal classifier;
\item if $\lambda=+\infty$, then the minimizer w.r.t. $\phi_1$ converges faster to the Bayes optimal classifier;
\item if $\lambda=0$, then the minimizer w.r.t. $\phi_2$ converges faster to the Bayes optimal classifier.
\end{itemize}
\begin{proof}
Following Proposition 1 and Lemma 3, we have,
\begin{equation}
\psi_i(\theta) = M_i\theta^{\alpha_i}+o(\theta^{\alpha_i})\quad \text{for}\quad  i=1,2 \quad \text{and}\quad 0<M_i<\infty
\end{equation}
Then, we get,
\begin{equation}
\lambda=\lim_{\theta\rightarrow 0+}{\frac{M_1\theta^{\alpha_1}+
o(\theta^{\alpha_1})}{M_2\theta^{\alpha_2}+o(\theta^{\alpha_2})}}=
\frac{M_1}{M_2}\lim_{\theta\rightarrow 0+}\theta^{\alpha_1-\alpha_2}
\end{equation}
Thus, we can conclude:
\begin{itemize}
\item for $\lambda=\frac{M_1}{M_2}\in (0,+\infty)$, then we have $\alpha_1=\alpha_2$ and $I_1=I_2$. Therefore, the minimizers w.r.t. $\phi_1$ and $\phi_2$ converge equally fast to the Bayes optimal classifier;
\item for $\lambda=+\infty$, we have $\alpha_1<\alpha_2$ and $I_1>I_2$. Thus the minimizer w.r.t. $\phi_1$ converges faster to the Bayes optimal classifier;
\item for $\lambda=0$, then we have $\alpha_1>\alpha_2$ and $I_1<I_2$, which means that the minimizer w.r.t. $\phi_2$ converges faster to the Bayes optimal classifier.
 \end{itemize}
 \end{proof}
\end{theorem}
Using the notion of intensity ratio, we can compare  the convergence rate to the Bayes optimal classifier for any two algorithms with Bayes-risk consistent loss functions without computing the consistency intensity.

We finish this section by introducing a scaling invariant property of the consistency intensity $I$, which is useful for comparing the convergence rate, e.g., when we scale the surrogate loss function $\phi(z)$ by $k_2\phi(k_1z)$, we get the same $I$ for $\phi(z)$ and $k_2\phi(k_1z)$.
\begin{theorem} \label{thm5}
For any constants $0<k_1,k_2<+\infty$, the loss $\tilde{\phi}(z)=k_2\phi(k_1z)$ have the same consistency intensity $I$ as that of $\phi(z)$, which means that the intensity of surrogate loss function is scaling invariant in terms of $\phi$.
\begin{proof}
Notice that $\psi(R(f_n)-R^{*})\leq R_{\phi}(f_n)-R_{\phi}^{*}$. If we scale $\phi$ as $\tilde{\phi}(z)=k_2\phi(z)$, the both sides of the inequality will be multiplied by $k_2$, which holds trivially. \par\noindent  We now consider $\tilde{\phi}(z)=\phi(k_1z)$. Observe that,
\begin{equation}
\begin{aligned}
&\tilde{C}^{*}(\eta)=\inf_{z\in\mathbb{R}}\eta \phi(k_1z)+(1-\eta)\phi(-k_1z)\\&=\inf_{k_1z\in\mathbb{R}}\eta \phi(k_1z)+(1-\eta)\phi(-k_1z) \\&=\inf_{z'\in\mathbb{R}}\eta \phi(z')+(1-\eta)\phi(-z') = C^{*}(\eta)
\end{aligned}
\end{equation}
where $z'=k_1z$. Then,
\begin{equation}
\tilde{\psi}(\theta)=\tilde{\phi}(0)-\tilde{C}^{*}(\eta)=\phi(0)-C^{*}(\eta)=\psi(\theta)~,
\end{equation}
which leads to the same $I$.
\end{proof}
\end{theorem}

Theorem \ref{thm5} has many applications. For example, the exponential loss in Adaboost is $\phi(z)=e^{-z}$. Then $\phi_k(z)=e^{-kz}$ for all constants $k>0$ must have the same consistency intensity as that of $\phi(z)$, which implies the minimizers of the corresponding empirical surrogate risks converge equally fast to the Bayes optimal classifier.

\subsection{$\text{SVM}_\Delta$: An Example of the Data-dependent Loss Modification Method}\label{app2}
In the previous sections, we have provided theorems that can measure the convergence rate from the expected risk $R(f_n)$ to the Bayes risk $R^*$ for many leaning algorithms using different surrogate loss functions. In fact, the notions of consistency intensity and conductivity can achieve something beyond that.  In this subsection, we propose a data-dependent loss modification method for SVM, that obtains a faster convergence rate for the bound of the excess risk $R(f_n)-R^*$ and thus makes the learning algorithm achieve a faster convergence rate to the Bayes optimal classifier.

We are familiar with the standard SVM that uses the hinge loss as a surrogate. Now, we modify the hinge loss as follows:
\begin{equation}
\phi(z)=max\{1-z,0\}^{1+\Delta} \quad \text{where} \quad 0<\Delta<+\infty ~.
\end{equation}
We have $\phi'(0)=-(1+\Delta)<0$, which means this modified hinge loss is also Bayes-risk consistent. Then, following the similar procedure as done for the above examples, we have,
\par\noindent
\begin{equation}
C(\eta)=\eta*max\{1-z,0\}^{1+\Delta}+(1-\eta)*max\{1+z,0\}^{1+\Delta}~.
\end{equation}
It's easy to verify that
\begin{equation}
z^{*}(\eta)=\frac{\eta^{\frac{1}{\Delta}}-(1-\eta)^{\frac{1}{\Delta}}}{\eta^{\frac{1}{\Delta}}+(1-\eta)^{\frac{1}{\Delta}}}~,
\end{equation}
and so
\begin{equation}
C^{*}(\eta)=\frac{2^{1+\Delta}\eta(1-\eta)}{[\eta^{\frac{1}{\Delta}}+(1-\eta)^{\frac{1}{\Delta}}]^{\Delta}}~.
\end{equation}
Thus,
\begin{equation}
\begin{aligned}
&\psi(\theta)=\phi(0)-C^{*}(\frac{1+\theta}{2})\\&=1-\frac{2^{\Delta}(1-\theta^2)}{[(1+\theta)^{\frac{1}{\Delta}}+(1-\theta)^{\frac{1}{\Delta}}]^{\Delta}}~.
\end{aligned}
\end{equation}
Using Maclaurin expansion, we have,
\begin{equation}
\psi(\theta)=\frac{1+\Delta}{2\Delta}\theta^2+o(\theta^2)~.
\end{equation}
Therefore, we have that intensity $I=\frac{1}{2}$ and conductivity $S=\sqrt{\frac{2\Delta}{1+\Delta}}$. According to Theorem \ref{thm2}, we know that, with a high probability, we have,
\begin{equation} \label{eq43}~
R(f_n)-R^*\leq\mathcal{O}\left(\frac{S}{n^{pI}}\right)=\mathcal{O}\left(\sqrt{\frac{2\Delta}{1+\Delta}*\frac{1}{n^p}}\right)~.
\end{equation}
From (\ref{eq43}), we find that a tighter bound can be obtained when $\Delta$ converges to zero fast. Here, we introduce the notion of data-dependent loss modification. That is, the modification parameter $\Delta$ is dependent on the sample size $n$.
For example, if $\Delta=\mathcal{O}(\frac{1}{n^2})$ and $p=\frac{1}{2}$, with a high probability, we can obtain a bound of order $\mathcal{O}\left(\frac{1}{n^{\frac{5}{4}}}\right)$, which, to our best knowledge, is the first bound faster than $\mathcal{O}\left(\frac{1}{n}\right)$ without the low noise assumption. Therefore, a tighter bound is obtained with our proposed method for $\text{SVM}$.

\section{Conclusions and Future Work} \label{sec5}
In this paper, we defined the notions of consistency intensity and conductivity for convex Bayes-risk consistent surrogate loss functions and proposed a general framework that determines the relationship between the convergence rate of the excess risk and the convergence rate of the excess $\phi$-risk. Our methods were used to compare the convergence rates to the Bayes optimal classifier for empirical minimizers of different classification algorithms. Moreover, we used the notions of consistency intensity and conductivity to guide modifying of surrogate loss functions so as to achieve a faster convergence rate.

In this work, we need the surrogate loss function to be convex and Bayes-risk consistent, which holds true for many different surrogate loss functions. However, sometimes we may encounter non-convex, but still Bayes-risk consistent loss functions. It is interesting to generalize the obtained results to the non-convex situation in the future. Besides,  in the future, we will apply the modified surrogate loss function to some real-world problems. Moreover, finding some other approaches that guide modifying existing surrogate losses to achieve a faster convergence rate to the Bayes optimal classifier is also quite worth exploring.

\bibliographystyle{apalike}
\bibliography{bare_jrnl.bib}

\begin{IEEEbiography}[{\includegraphics[width=1in,height=1.25in,clip,keepaspectratio]{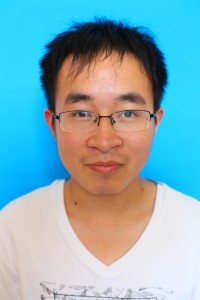}}]{Jingwei Zhang}
received the MPhil degree in computer science from the University of Sydney in 2019 and B.E. degree in electronics engineering and information science from the University of Science and Technology of China, in 2017. He is currently pursuing the Ph.D. degree in computer science from the Hong Kong University of Science and Technology. His research interests include machine learning and theory of deep learning.
\end{IEEEbiography}

\begin{IEEEbiography}[{\includegraphics[width=1in,height=1.25in,clip,keepaspectratio]{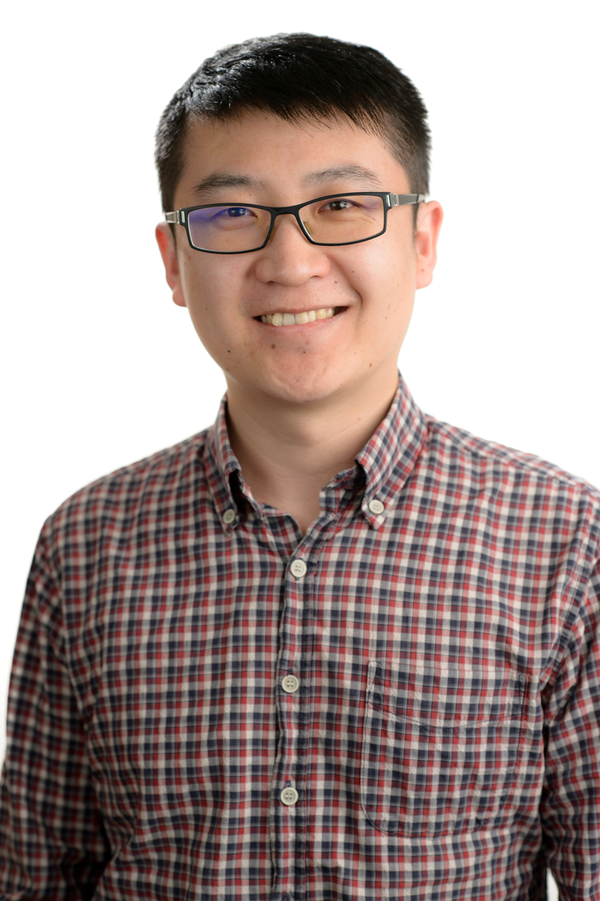}}]{Tongliang Liu}
received the B.E. degree in electronic engineering and information science from the University of Science and Technology of Chi- na, Hefei, China, in 2012, and the Ph.D. degree from the University of Technology Sydney, Sydney, Australia, in 2016. He was a visiting Ph.D. student with the Barcelona Graduate School of Economics and the Department of Economics, Pompeu Fabra University, for six months. He is currently a Lecturer with the School of Computer Science and the Faculty of Engineering and Information Technologies, the University of Sydney. He has authored or co-authored over ten research papers, including the IEEE T- PAMI, T-NNLS, T-IP, NECO, ICML, KDD, IJCAI, and AAAI. His research interests include statistical learning theory, computer vision, and optimization. He received the Best Paper Award in the IEEE International Conference on Information Science and Technology 2014.
\end{IEEEbiography}


\begin{IEEEbiography}[{\includegraphics[width=1in,height=1.25in,clip,keepaspectratio]{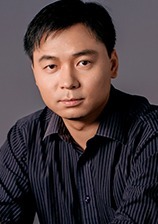}}]{Dacheng Tao}
(F'15) is currently a Professor of Computer Science with the School of Computer Science and the Faculty of Engineering and Information Technologies, the University of Sydney. He mainly applies statistics and mathematics to artificial intelligence and data science. His research interests spread across computer vision, data science, image processing, machine learning, and video surveillance. His research results have expounded in one monograph and over 200 publications at prestigious journals and
prominent conferences, such as the IEEE T-PAMI, T-NNLS, T-IP, JMLR, IJCV, NIPS, ICML, CVPR, ICCV, ECCV, AISTATS, ICDM, and ACM SIGKDD, with several best paper awards, such as the best theory/algorithm paper runner up award in the IEEE ICDM07, the best student paper award in the IEEE ICDM13, and the 2014 ICDM 10-year highest-impact paper award. He is a fellow of the IEEE, OSA, IAPR, and SPIE. He received the 2015 Australian Scopus-Eureka Prize, the 2015 ACS Gold Disruptor Award, and the 2015 UTS Vice-Chancellors Medal for Exceptional Research.
\end{IEEEbiography}





\end{document}